# UTILIZATION OF A LEXICON FOR SPELLING CORRECTION IN MODERN GREEK

A. Vagelatos, T. Triantopoulou, C. Tsalidis, D. Christodoulakis
Computer Technology Institute & Computer Engineering Dept. University of Patras

**Abstract**
In this paper we present an interactive spelling correction system for Modern Greek (M.G.). The entire system is based on a morphological lexicon. Emphasis is given to the development of the lexicon, especially as far as storage economy, speed efficiency and dictionary coverage are concerned. Extensive research was conducted from both the computer engineering and linguistic fields, in order to describe inflectional morphology as economically as possible.

**Introduction**
Three years ago, we undertook a project, called *Intralex* (fully supported by the Greek telecom. industry INTRACOM), aiming at the development of an interactive spelling checking/correction system for M.G., to be based on as a complete as possible *lexicon*, designed to "run" on MS-DOS based computers. Quality performance of such an interactive system was a major task; great response time would result in an inconvenient system. Our main concerns focused on:
1. *Storage economy*. It would be absurd to store each word plus all its forms directly in the computer, because M.G. is a highly inflectional language thus vast amounts of storage would be required. Storage economy was attained by devising a system to code M.G. inflections and marked stress. This was successfully achieved through an extensive analysis of M.G. morphology.
2. *Speed efficiency*. This is a very important factor in the design of interactive systems. Speed efficiency was achieved by using sophisticated data structures for the storage of the dictionaries.
3. *Dictionary coverage*. Since the basis of any spelling checker is its dictionary, the quality of the entire system is analogous to its dictionary coverage. Therefore, we aimed at providing thorough dictionary coverage.
4. *Optimum correction schema*. Apart from the checking, the correction schema is equally important for the user of such a system. In order to reach an optimum correction mechanism, we adopted a combination of correction techniques, keeping in mind the special linguistic characteristics of M.G.

**Modern Greek morphology**
Aiming at the development of a system that would process M.G. at the morphological level, an analysis of M.G. morphology was carried out[2,6,9,11,14,15]. This analysis, gave the guidelines and the directions thus providing the solutions to the particular problems that were encountered.

The main characteristics of M.G. morphology can be summarized as:
- A complex inflectional system. For example, for the M.G. masculine nouns ending in -ος /os/, there are six different inflections and for the present tense of active voice of verbs ending in -ω /o/ there are seven different inflections.
- The existence of marked stress: Words in M.G. are stressed in either the final, penultimate or antepenultimate positions; e.g., εδώ /edo/, τώρα /tora/, κάποτε /kapote/, respectively.
- A "graphematic" spelling system consisting of single graphemes; e.g., α=/a/, ε=/e/, compound graphemes; e.g., αι=/ε/ and grapheme equivalents; e.g., ι,η,υ=/e/, ω,o=/o/.
- The existence of numerous characteristics carried over from Ancient Greek. The use of old and new forms, which give rise to an endless linguistic debate in regard to whether both forms should continue to be used (accepted). This is illustrated in the first person plural of the verb "παίζ-ω" /pεzo/ where two forms exist and are acceptable: "παί-ζου-με" /pε-zu-mε/ (new) and "παί-ζο-με" /pε-zo-mε/ (old).

M.G. word classification incorporates two basic categories: The *inflected* and the *non-inflected*. Our main task was to study inflected M.G. words in order to describe their declinations and conjugations as economically as possible.
All inflected words were given the following morphological description:

  WORD = [PREFIX] + STEM + [INFIX] + INFLECTION(S)

where: STEM and INFLECTION(S) are necessary features for the derivation of all possible forms, PREFIX and INFIX are not always present and thus appear bracketed "[ ]" in the above description; e.g.,

ξανα-γράφ-τηκ-α = PREFIX(ξανα) + STEM(γράφ) +
/ksana-γraf-tik-a/    INFIX(τηκ)+INFLECTION(α)

έ-γραφ-α         =PREFIX(ε) + STEM(γραφ) +
/ε-γraf-a/              INFLECTION(α)

### Use of a description language to Code the inflectional morphology and marked stress of M.G.

Building a morphological lexicon to be the base of a spelling correction system, is not an easy task especially when:
- one has to deal with a highly inflectional language with marked stress,
- one has to make it as complete as possible, since the target in this case was a commercial product and not just a prototype.

With the above constraints, serious consideration had to be given to *speed efficiency* and *disk space*. Thus, our aim was to find a way to describe, most economically such enormous information to the fullest, while, and at the same time avoid redundancy.

Keeping the above factors in mind, we decided on a scheme consisting of a *context-free description language* which we called -*Greek Word Description Language, (GWDL)*- describing both inflectional morphology and stress of M.G.[16]. Thus, given the stem of every declinable word, the appropriate set of rules are attached, making possible the production of all valid forms for the particular word. Every such combination is included in the lexicon, forming a lexicon entry. This way all words with the same root, are stored as a single lexical item along with rules for allowable inflection and stress, thus saving vast amounts of storage.

### The Greek Word Description Language (GWDL)

The *Greek Word Description Language* (GWDL), as we developed it, comprises a set of rules that code the declinable part of inflected words. It utilizes 290 rules. Using GWDL, only the stem of each word followed by a rule (or a set of rules) representing possible endings, are stored in the lexicon. It must be mentioned here, that we had to deal with a lexicon of nearly 90.000 stems where for almost each stem, a set of rules was required.

GWDL which forms an inflection generator for M.G., thus coded:
- Inflection: suffix endings for gender, number, case, person, etc. Inflection is denoted by inflection rules, e.g., #OUSOSa= ος|ου|ο|ε|οι|ων|ους.
- Stress: in general, words in M.G. may be stressed on the last three syllables only. Marked stress must not occur in single syllable words. Stress is denoted by stress rules, e.g.,
  
  !a1=(1). (Stress in final position.)
  !a2=(2). (Stress in penultimate position.)
  !a6=(3). (Stress in antepenultimate position.)
- Infix: infix is denoted by combining rules which consist of infix and inflection/stress rules, e.g., Past Tense of Active Voice:
  
  $AEURA1= χ #PAESY !a6. , where:
  
  (χ=infix, #PAESY=inflection rule, !a6=stress rule).

The general principles underlying GWDL are as follows:
- All stems are left uncoded but are syllabificated. Syllabification was necessary to enable handling of marked stress.
- At the end of non-inflected M.G. words, only stress need be denoted, e.g., κα-που!a2 = κά-που, where !a2 is the GWDL stress rule which represents penultimate stress.
- At the end of inflected M.G. words combining rules consisting of inflection-stress rules or infix-inflection-stress rules can be applied, e.g.,
  
  προ-ο-δ[$OUSOS7].

### BNF description of the GWDL
The description of the GWDL in BNF notation is:

```
lexicon_file   ::= ['%%'] definition_part '%%' words_part
definition_part ::= { definition }
definition     ::= stress_def | inflection_def | form_def
stress_def     ::= STRESSV '=' stress
stress         ::= '(' NUMBER { ',' NUMBER } ')'
inflection_def ::= INFLECTIONV '=' inflection
inflection     ::= '[' SUFFIX { '|' SUFFIX } [ '|' ] ']'
form_def       ::= FORMV '=' form { | form }
form           ::= [INFIX] (INFLECTIONV | inflection)
                   ( STRESSV | stress)
words_part     ::= { word }
word ::= [ STEM ] '[' form { | form } ']' '.' | STEM stress'.'
```

To exemplify the above BNF description we must mention that the source lexicon files have two parts: The definition part where the rules (GWDL) are defined and the word part where the lexicon entries reside.

### Examples of coded words
Following are two examples of coded words, together with the word forms that can be produced from them:

-- noun: "πρόοδος" /proodos/ (progress)

| *Lexicon Entry* | *Produced Inflected Forms* |
|---|---|
| προ-ο-δ[$OUSOS7]. | πρόοδος |
| | προόδου |
| | πρόοδο |
| | πρόοδοι |
| | προόδων |
| | προόδους |

where,
$OUSOS7   = #OUSOSb ! a14.
#OUSOSb   = ος | ου | ο | οι | ων | ους.
!a14      = (3, 2, 3, 3, 2).

-- verb: "αγαπώ" /αγαpo/ (love)

| *Lexicon Entry* | *Produced Inflected Forms* |
|---|---|
| α-γα-π [$ENEAO| | αγαπώ, αγαπάς, αγαπά, |
| | αγαπάμε, αγαπούμε, αγαπάτε, |
| | αγαπάν, αγαπάνε, αγαπούν. |

$PAEF1]. αγαπούσα, αγαπούσες,
αγαπούσε, αγαπούσαμε,
αγαπούσατε, αγαπούσαν
αγαπούσανε.

where,
$ ENEAO = #ENEAO !b1.
#ENEAO= ω|ας|α|ουν|αν|αμε|ουμε|ατε|ανε|ουνε.
!b1 = (2,2,2,2,3).
$PAEF1= ουσ # PAE !b1.
#PAE = α|ες|ε|αν|αμε|ατε|ανε.

### Lexicon development
#### Overall principles

Our lexicon contains 90.000 entries. The possible word forms produced from these entries have been calculated to exceed one million. With each stem the appropriate production rule(s) are related, in order to produce all the distinct yet acceptable M.G. word forms.

The primary storage mechanism used to access words in the Lexicon is the "Compressed Trie"[7]. It was established that this data structure was the most appropriate storage method. This method enables efficient search and occupies less disk storage. The Compressed Trie is used as an index to the database of the words. This structure is relatively small (about 700Kb) compared to data needed to represent the entire lexicon. Thus, we can load a big part of it (or the whole, if the computer has enough memory) into main memory. The Compressed Trie contains the part of a word's stem necessary to distinguish this word from stems of all other words having the same prefix.

The Compressed Trie is also very useful in the *correction*, where we normally search hundreds of alternative words. If there is not an acceptable prefix in the Compressed Trie for the alternative word we stop the search for this alternative and continue with the next.

The number of infixes and inflections used is very small in comparison to the number of words. For the 90.000 stems of the Lexicon database, there are about 400 distinct infixes and 200 distinct inflections which means that these infixes and inflections are used very frequently in order to cover the 90.000 stems. Consequently, it is more efficient to keep them in main memory stored in a "Symbol Table".

The actual data of words are stored in a file on disk. We access the position where the data for a particular word exist through the Compressed Trie Index. If the compressed trie is in main memory then we approximately make one disk access per search for words located on the disk dictionary.

#### Supporting tools

The development of the lexicon was a long and tedious process in which many difficulties were encountered. The main difficulty was to successfully attach the appropriate rules to each stem, so as to derive all possible word forms while avoiding the production of unacceptable forms, as well as avoiding redundancy and overlapping.

In order to simplify this process an environment was built consisting of the following tools:

**1.** A program capable of automatically producing lexicon entries (stem+ inflection and stress rules) of the 90.000 words that had been stored in computer readable form. The program was able to cut-off the ending of each word, and decide based on this ending, the word category (e.g., verb in past tense of active voice), so that, the appropriate rules could be attached. This process resulted in about 85% accuracy, although rules that resulted in some unacceptable/meaningless forms were attached at times. After this preprocessing the linguists undertook the task of validating the lexicon entries manually.

**2.** A *syntax oriented editor*. This editor assisted in two ways: First it helped validate the syntax rules for each entry (whether or not the rules were syntactically correct). Second, having the ability to produce the inflected forms of each entry, the linguists were able to check the "correctness"/acceptability of each entry and also the completeness of the GWDL. This way, additional rules were provided so as to result in a description language that would be complete.

**3.** *Mkdict* (make dictionary), is a program which was developed in order to construct the dictionaries. It takes as input the lexicon files (which contain the word forms as described above) and the set of rules, and constructs the final dictionaries.

### Spelling Correction Methodology
#### Error types

Spelling errors can be categorized into the following types[3,5,10]:

**a)** *Orthographical errors*. These are cognitive errors consisting of the substitution of a deviant spelling for a correct one when the author either simply doesn't know the correct spelling for a word, forgot it or misconceived it.

The important characteristic of orthographical errors is that they generally result in a string which is phonologically identical or very similar to the correct one. As a consequence orthographical errors depend on the correspondence between spelling and pronunciation of a particular language.

**b)** *Typographical errors*, are motoric errors, caused by hitting the wrong sequence of keys. Hence, their characteristics depend on the use of a particular keyboard rather than a particular language. They are further categorized as:

1) deletion errors, e.g., "πρόραμμα" instead of "πρόγραμμα" (program),
2) insertion errors, e.g., "πρόγφραμμα",
3) substitution errors, e.g., "πρόγταμμα",

4) transposition errors, e.g., "πρόργαμμα".

**Correction schema**
The above error categorizations generally apply to every natural language. More specifically, for M.G. it was applied as follows:

In the case of *orthographical errors*, M.G. has the following homophonous sets of vowels or vowel blends:
1) ε - αι :/ε/, 2) o - ω :/o/, 3) η - ι - υ - ει - οι :/i/
Apart from the above vowels, the following allophone combinations of vowels and consonants exist:
1) χθ - χτ :/χθ - χt/,   2) φθ - φτ :/fθ - ft/,
3) σθ - στ :/sθ - st/,   4) αυ - αβ :/av/,
5) ψ - πσ :/ps/,   6) ξ - κσ :/ks/.
Thus: Given an incorrect word we first attempt to find out if it has an orthographical error. Using the above sets of letters, we produce all the possible strings (we use the term "string" because after the substitution, the outcome does not always result in a valid word) from the incorrect word by substituting each of the vowels of a set with another vowel of the same set. The produced strings are then looked up in the lexicon for a match. Thus, each string produced has to be matched to one word in the lexicon to become a possible alternative for the incorrect word.

In the case of *typographical errors* we use the so called "error reversal" method[10] with some modifications for better efficiency. The error reversal method is based on the idea that *from an incorrect word the correct word can be produced, if we apply the error type in "reverse"*. For instance if we have an insertion error e.g., "πρόγφραμμα" instead of "πρόγραμμα", by deleting the incorrect letter we can produce the correct word. As a matter of fact, in the above example we have to delete all the letters of the word, one by one, and look up the resulting strings in the lexicon to find the valid matches. Even more, because the error type cannot be predicted, we have to apply all the error rules in reverse, to find the possible correct word.

It ought to be mentioned that especially in the process of reversing the deletion error (which in fact is the insertion of possible letters), yields to an enormous number of possible strings which have to be looked up in the lexicon; e.g., trying to reverse the deletion error in the word "πρόγαμμα", we must insert, one by one, all the letters of the Greek alphabet (24) starting from the position before the first letter of the word to the position after the last letter of the word (24 X 9 = 216 searches in the lexicon). To reduce the number of the produced words we use *trigrams*. Trigrams are the valid three letter strings that appear in any word in a language. During the insertion we are careful not to produce words with invalid trigrams, thus incorrect ones.

Finally in M.G. we were faced with yet another error type namely, **stress position errors**, e.g., "κέφαλι" /kεfali/ (head) instead of "κεφάλι". The correction of this error type is based on the lexicon structure. As we mentioned earlier, the words are stored in the lexicon without stress; stress follows in code form in the rule part of the entry (e.g., !a1). This way every word is searched without the stress and as soon as an entry has been matched, stress is added. If the stress is in a different position, then there is probably a stress position error and the word found is suggested as an alternative.

**Overall System Integration**
The overall system design was developed with emphasis on speed, efficiency and user friendliness. More specifically, the system is based on three different dictionaries:
- The *Memory-Resident* dictionary for storage of the most commonly used M.G. words. This dictionary contains about 800 words. These words were collected after a statistical analysis of a great number of M.G. texts. We used a "compressed-trie" for indexing the roots of the words.
- The *Main* dictionary (described earlier), residing on the disk, where the main part of the M.G. dictionary is stored. The dictionary's access time depends only on the *length* of a word and not on the number of the words in the dictionary.
- The *User* dictionary where user-specific words (e.g., terminology), are stored by the user. For the implementation of this dictionary we used a "hash table".

The entire system is an interactive program which takes as input an ASCII file, or a file created from one of the four word-processors with which it is compatible, and executes the spelling check. As soon as an "incorrect" word has been found (a word that doesn't exist in the dictionaries), the user is prompted to select from the following choices:
**Skip** this word.
**Edit** the word.
**Store** the word in the user's dictionary.
**Correct**. (Gives possible corrections by executing the correction algorithms.)
**Exit** the checking phase.

If the choice "correct" is made, the correction method is initiated and possible corrections are provided. If any of these corrections result in valid words, they are suggested to the user as alternatives. The user can then select the appropriate alternative to replace the misspelled word.

The performance of the existing system depends on the computer type and on the type of text. On a 386/33MHz computer with 2Mb extended memory the checking rate varies from 2000 words/sec (when most of the checking words are in the memory dictionary) to 60 words/sec

(when most of the checking words are in the main dictionary).

**Conclusions**

In the above paragraphs we presented the experience that was gained during the development of a spelling correction system. We followed some well known techniques but with certain adaptations and modifications in order to handle the peculiarities of M.G.
More specifically, we were able to achieve:
a) The coding of the inflectional morphology, which seems to be the most efficient method for lexicon development of highly inflectional languages.
b) The development of supporting tools for the construction of the lexicon. The existence of such tools proved to be necessary for the construction and maintenance of wide coverage lexicons.
c) The development of an appropriate correction schema.
Apart from the above, and since the system was to become a commercial product, great effort was put into attaining high vocabulary coverage. This is indeed one of the advantages of the system as we had predicted and as was verified by feedback.

**Future Directions**

The Intralex spelling correction system has already become a commercial product in the form of an interactive program. Furthermore, it has been approved by Microsoft corp., as the Greek spelling module that will be marketed with the Greek versions of their products. Our future plans aim at expanding:
- the GWDL, for full M.G. derivational morphology coverage.
- the Intralex lexicon entries so that syntactic as well as semantic information be included.

Such an expanded lexicon could be the basis for the construction of a reusable and multi-purpose grammar, as well as an efficient parser for the syntactic and semantic analysis of Modern Greek. The resulting grammar is intended as a:
- starting point for further linguistic research, incorporating semantics and the formal properties of written text data,
- basis of the construction of customized Natural Language interfaces and syntax-directed full-text retrieval systems,
- basis for the construction of syntactic text processing tools,
- platform for research in comparative linguistics and mechanical translation.

**Acknowledgments**
We thank all team members, and especially M. Stamison-Atmatzidi (EFL Instructor/Computational Linguistics researcher), whose guidelines, valuable comments, proofreading and fine tuning of English usage were indispensable, and G. Moutsos (Computer Engineer), for his help on implementing the S/W.

E-mail address: vagelat/dtriant/tsalidis/dxri@cti.gr